\documentclass[conference]{IEEEtran}

\usepackage[pdftex]{graphicx}
\usepackage{booktabs, multirow}
\usepackage{enumitem}
\usepackage{subcaption}
\usepackage{amsmath, amssymb, amsfonts, bm, bbm}
\usepackage{hyperref}
\usepackage{pifont}% http://ctan.org/pkg/pifont
\newcommand{\cmark}{\ding{51}}%
\newcommand{\xmark}{\ding{55}}%

\newcommand{\R}{\mathbb{R}}

\newcommand{\D}{\mathcal{D}}
\newcommand{\U}{\mathcal{U}}
\newcommand{\I}{\mathcal{I}}
\newcommand{\calL}{\mathcal{L}}
\newcommand{\bfL}{L}
\newcommand{\Layernorm}{\mathrm{LayerNorm}}
\newcommand{\Dropout}{\mathrm{Dropout}}
\newcommand{\Sigmoid}{\mathrm{Sigmoid}}
\newcommand{\ReLU}{\mathrm{ReLU}}
\newcommand{\SA}{\mathrm{SA}}
\newcommand{\Attention}{\mathrm{Attention}}
\newcommand{\Softmax}{\mathrm{Softmax}}
\newcommand{\FF}{\mathrm{FF}}
\newcommand{\IndicatorUIL}{\mathbbm{1}_{(u,i,l) \in \D^{+}}}
\newcommand{\IndicatorUILGNeg}{\mathbbm{1}_{(u,i,l) \in \D_G^{-}}}
\newcommand{\IndicatorUILSSNNeg}{\mathbbm{1}_{(u,i,l) \in \D_{SSN}^{-}}}
\newcommand{\embeddim}{\bar{d}}

\begin{document}
\title{HyperTeNet: Hypergraph and Transformer-based Neural Network for Personalized List Continuation}

\author{
    \IEEEauthorblockN {
        Vijaikumar M.
    }
    \IEEEauthorblockA{
        \textit{Department of CSA} \\
        \textit{Indian Institute of Science} \\
        Bangalore, India \\
        vijaikumar@iisc.ac.in
    }
    \and
\IEEEauthorblockN {
        Deepesh Hada
    }
    \IEEEauthorblockA{
        \textit{Department of CSA} \\
        \textit{Indian Institute of Science} \\
        Bangalore, India \\
        deepeshhada@iisc.ac.in
    }
    \and
    \IEEEauthorblockN {
        Shirish Shevade
    }
    \IEEEauthorblockA{
        \textit{Department of CSA} \\
        \textit{Indian Institute of Science} \\
        Bangalore, India \\
        shirish@iisc.ac.in
    }
}

\maketitle

\begin{abstract}
The personalized list continuation (PLC) task is to curate the next items to user-generated lists (ordered sequence of items) in a personalized way. The main challenge in this task is understanding the ternary relationships among the interacting entities (users, items, and lists) that the existing works do not consider. Further, they do not take into account the multi-hop relationships among entities of the same type. In addition, capturing the sequential information amongst the items already present in the list also plays a vital role in determining the next relevant items that get curated.

In this work, we propose HyperTeNet -- a self-attention hypergraph and Transformer-based neural network architecture for the personalized list continuation task to address the challenges mentioned above. We use graph convolutions to learn the multi-hop relationship among the entities of the same type and leverage a self-attention-based hypergraph neural network to learn the ternary relationships among the interacting entities via hyperlink prediction in a 3-uniform hypergraph. Further, the entity embeddings are shared with a Transformer-based architecture and are learned through an alternating optimization procedure. As a result, this network also learns the sequential information needed to curate the next items to be added to the list. Experimental results demonstrate that HyperTeNet significantly outperforms the other state-of-the-art models on real-world datasets. Our implementation and datasets are available online\footnote{\url{https://github.com/mvijaikumar/HyperTeNet}}.
\end{abstract}

\IEEEpeerreviewmaketitle

\section{Introduction}
The personalized item list continuation (or personalized list continuation (PLC)) task is to recommend a potential sequence of items to user-generated item lists in a personalized way. In recent years, the personalized list continuation problem has been receiving significant attention due to its application in various online services. For example, Spotify, a music streaming platform, requires automatically identifying a potential sequence of tracks (songs) to user-generated playlists. Goodreads requires a recommender system that automatically recommends a list of books for user-generated themes.

In recommender systems, most of the tasks can be modeled as a link prediction problem. For example, recommending top-N items for each user can be modeled as predicting links between users and items. However, the personalized list continuation task requires link prediction that goes beyond pairwise relations. For example, consider a user who has two playlists having different genres, and let us say she does not interact with songs apart from these genres. Let us assume that the first playlist primarily consists of progressive rock songs while the second comprises Blues songs. Good recommendations for the first playlist would consist of songs by, for instance, \textit{Pink Floyd}, and \textit{Porcupine Tree}. For the Blues playlist, good recommendations would be songs by the legends \textit{Gary Moore} and \textit{B. B. King}. Since the user has not interacted with a song from other genres, recommending a song (for example, \textit{Coffin Dance}) from another genre (\textit{EDM}) to any playlists could hamper the user experience. Recommending songs to a user under a specific playlist must account for her interests and the playlist. So, we are interested in predicting a link that concurrently connects user, music, and playlist (user-music-playlist). However, modeling the above problem using only pairwise link prediction incurs a loss of information. For instance, a link between a user and a song disregards information about the playlist. Similarly, predicting a link between a playlist and a song snubs the information about the user for whom the music is recommended.

At the same time, items are added to user-generated lists \textit{sequentially}, and capturing this sequential information is essential to understand users' dynamically changing behavior. In the earlier example, let us assume that the user starts liking songs having hard and alternative rock genres. This may lead her to add these songs to one of the existing lists or create a new playlist altogether. Besides, when defining a playlist, we argue that not all the tracks contribute equally, \textit{i.e.}, some tracks may have more influence in the construction of the playlist and in the addition of new tracks into it than the others. This corresponds to the user's general preference. Hence, the model should not only be able to capture the sequential information but also be able to learn the influence of tracks on defining lists and newly added tracks.

\begin{table}[bt]
    \Large
    \centering
    \caption{Summary of different models proposed for personalized list continuation task.}
    \label{tab:com_models}
    % \begin{tabular}{p{4cm} p{1.5cm} p{1.5cm}p{1.5cm}p{2.5cm} p{3cm} p{2cm} }
    \begingroup
    \renewcommand*{\arraystretch}{1.3}
    \resizebox{\columnwidth}{!}{
    \begin{tabular}{lccccc}
        \toprule
        \multirow{2}{*}[-1pt]{Property} & \multirow{2}{*}[-1pt]{SASRec~\cite{kang2018self}} & \multirow{2}{*}[-1pt]{CAR~\cite{he2020consistency}} & \multirow{2}{*}[-1pt]{Caser~\cite{tang2018personalized}} & \multirow{2}{*}[-1pt]{AMASR~\cite{tran2019adversarial}} & HyperTeNet \\ 
        & & & & & (this paper) \\
        \midrule
        Sequence-aware & \cmark & \cmark & \cmark & \xmark & \cmark \\
        Personalized & \cmark & \cmark & \cmark & \cmark & \cmark \\
        Attention-based & \cmark & \cmark & \xmark & \cmark & \cmark \\
        Higher-order graph structure & \xmark & \xmark & \xmark & \xmark & \cmark \\
        Ternary relationships & \xmark & \xmark & \xmark & \xmark & \cmark \\
        \bottomrule
    \end{tabular}
    }
    \endgroup
\end{table}

Although models proposed for sequential recommendation can be used for the personalized list continuation task, they have some drawbacks. For example, Convolutional Sequence Embedding Recommendation Model (Caser) \cite{tang2018personalized} embeds a sequence of items into an image and learns the user's dynamic preferences using convolutional filters. Caser, however, does not accommodate a user's long-term behavior and focuses just on her dynamically changing behavior. Capturing the long-term static user behavior is essential for list continuation to understand the list theme. The Self-Attentive Sequential Recommendation (SASRec) \cite{kang2018self} model uses attention mechanism to capture long-term behavior by attentively selecting few actions for the next item recommendation. Hence, it achieves better results than Caser, justifying the importance of capturing the static behavior in lists. The Adversarial Mahalanobis distance-based Attentive Song Recommender (AMASR) \cite{tran2019adversarial}, specifically proposed for personalized song-playlist continuation, leverages attention mechanism and the Mahalanobis distance-based metric learning approach. However, it does not capture the sequential information in lists, indicating that the model cannot effectively understand the dynamically changing user behaviors. Consistency-aware and Attention-based Recommender (CAR) \cite{he2020consistency} utilizes an attention and gating mechanism to learn the consistency strength of the lists for item list continuation, capturing static and dynamic user behaviors. Nevertheless, none of the above models explicitly model the ternary relationship among the interacting entities (users, items, and lists). Also, they do not capture the relationships among the same types of entities, which is essential for the collaborative filtering setup and can be captured using the graph structure of recommendation datasets \cite{gnn_GraphCMC}. \autoref{tab:com_models} summarizes the different models proposed for personalized list continuation based on the properties they possess. The proposed model, HyperTeNet, not only exhibits all these properties but also outperforms these models for the personalized list continuation task on real-world datasets.

To capture the ternary relationships among users, items, and lists, HyperTeNet models the problem as a hyperlink prediction problem in a hypergraph whose nodes are users, items, and lists. By employing a hypergraph self-attention network \cite{zhang2020hypersagnn} that directly predicts the hyperlinks amongst users, items, and lists, we demonstrate that it is possible to recommend new items to an already existing user list or recommend a new list with new themes to the users. Further, effective use of the self-attention mechanism to capture sequential information and identify influential items in the lists is also shown. Experimental results on different real-world datasets demonstrate the effectiveness of our model against state-of-the-art models.

% To capture the relationship among users, items, and lists, HyperTeNet models this problem as a hyperlink prediction problem. We first construct $K$-NN graphs for users, items and lists using user-item, user-list, and list-item interactions. Then, we leverage Graph Neural Networks \cite{kipf2017semi} to incorporate neighborhood information to learn the representations for these entities. We then employ a hypergraph self-attention network \cite{zhang2020hypersagnn} that directly predicts the hyperlink amongst users, items, and lists. For example, in this context, a new hyperlink may denote (i) recommendation of new music to already existing playlist of the user; or (ii) recommendation of a new playlist with new music to the users. Further, to capture the sequential information as well as identifying influential items in the list for suggesting the next items, inspired by Transformer~\cite{vaswani2017attention} architecture, we leverage the self-attention mechanism.

To summarize, our work makes the following contributions:
\begin{itemize}
    \item We propose an end-to-end neural network, HyperTeNet, for the task of personalized list continuation (PLC) that \\
    (i) considers the generalized multi-hop relationships between homogeneous entities; \\
    (ii) captures the ternary relationship among users, items, and lists by constructing a 3-uniform hypergraph and modeling the interacting entities as a hyperedge; and \\
    (iii) acknowledges the dynamically changing user behavior for personalization.
    \item We empirically show that HyperTeNet significantly outperforms the current state-of-the-art baseline models on real-world datasets.
    \item Ablation studies are performed to demonstrate the effectiveness of HyperTeNet's sub-models.
    \item We study the impact of HyperTeNet's hyperparameters on the performance.
\end{itemize}

\section{Related Work}
Personalized ranking provides customers with item recommendations from a ranked list of items. An influential model, Bayesian Personalized Ranking (BPR) \cite{rendle2009bpr}, optimizes a pair-wise loss function for top-N item recommendation. A class of models does user-item Matrix Factorization (MF) \cite{general_nn_MF} to find latent space representations of users and items. With the arrival of deep learning, neural networks are used for collaborative filtering to generate strong recommendations \cite{he2017neural, general_nn_collaborativeDL, kim2016convolutional}. Neural Matrix Factorization (NeuMF) \cite{he2017neural} improves upon MF by modeling the user-item interaction with a neural network instead of a simple inner product. We refer the readers to the survey by Zhang et al. \cite{zhang2019deep} for a comprehensive investigation on deep learning-based recommender systems.
    
    \subsection{Hypergraph Neural Networks}

    Recommendation datasets have a natural interpretation in the form of graphs, as data can be modeled as relations between users and items \cite{gnn_GraphCMC}. Graph Neural Networks (GNNs) \cite{kipf2017semi} are much more flexible and conveniently model multi-hop relationships from user-item interactions. However, a relation can span more than two entities in many cases, as in paper-authorship and list continuation datasets, motivating the use of hypergraphs instead of graphs. Traditional hypergraph-based models decompose hyperedges into pairwise relationships. However, applying graph convolutions on hypergraphs may not be straightforward and may require building a simple graph instead. For example, a significant bulk of these works relies on clique expansion \cite{hgnn_spectral_learning, hgnn_cluster_classify, hgnn_DHNE}, which replaces each hyperedge with a clique on those vertices. Zien et al. \cite{hgnn_spectral_partioning} use a bipartite graph model (star expansion) to produce a graph having an edge between a hypernode $v$ and a hyperedge $e$ if $v \in e$. Some recent GNN-based models \cite{hgnn_HyperGCN, hgnn_hgnn, hgnn_hgcn_hgat} aim to generalize convolutions and attention mechanisms from graphs to hypergraphs. However, these models are primarily used for semi-supervised node classification tasks, where hypernode features are known and cannot be used to predict hyperedges directly. Hyper-SAGNN \cite{zhang2020hypersagnn} uses a self-attention-based hypergraph neural network and addresses this issue. The model learns node embeddings and can predict hyperedges in heterogeneous hypergraphs. Our work, HyperTeNet, employs an architecture motivated by Hyper-SAGNN to learn ternary relationships among users, items, and lists.

    \subsection{Sequential Recommendations}
    The main aim in providing sequential recommendations is to learn a sequence representation that mirrors the user's dynamic preference. Traditional methods \cite{seq_fuseMC, rendle2010factorizing} utilize Markov Chains (MC) to capture the item-to-item transitions in the sequence. Unfortunately, the Markov assumption is pretty strong, as it completely ignores a major chunk of the sequence. With the advent of deep learning, some models use Recurrent Neural Networks (RNN) to learn patterns in the sequence \cite{seq_rnn_session, seq_improved_rnn}. Caser \cite{tang2018personalized} takes a different approach by capturing high-order Markov Chains by applying convolutional filters on the most recent items. SASRec \cite{kang2018self} and BERT4Rec \cite{seq_BERT4Rec} leverage the self-attention mechanism to model item interactions within a sequence, inspired by the Transformer's \cite{vaswani2017attention} better long-range sequence modeling ability.
    
    \subsection{Personalized List Continuation}
    Sequential recommender systems seek to predict the next items that are likely to interact with a user based on her recent interaction history. In contrast, personalized list continuation systems curate the next items in the user's lists based on the user's general preferences, her dynamically changing behavior, and the list theme. A significant amount of previous efforts focus on personalized song-based playlist continuation. These approaches are known to use content features alongside collaborative filtering \cite{alc_context-aware, alc_coherent_continuation, alc_gaussian_prior, alc_two_stage_model, alc_approach_analysis, alc_TrailMix} to predict the next items in a given list. However, these approaches do not consider general lists, viz. book-based (Goodreads), answer-based (Zhihu), video-based (YouTube), and a multitude of other types of user-generated playlists. Most of these methods also implicitly assume the playlists to be consistent, \textit{i.e.}, all the items within the list belong to similar genres. Consistency-aware and Attention-based Recommender (CAR) \cite{he2020consistency} introduces the global preferences of users and a consistency-aware gating mechanism to capture heterogeneity in item consistency for list continuation.
    
    The research gap in the current literature exists in capturing ternary relationships among users, items, and lists. Also, a consolidated approach combining the multi-hop relationships between nodes in graphs, learning ternary relationships among different entity types, and capturing sequential information within lists seems missing. HyperTeNet is a unified framework for personalized automatic list continuation which aims to address these gaps. We discuss it next.

\section{The Proposed Model}
\noindent
\textbf{Personalized List Continuation.}
Let $n_U$, $n_I$, and $n_L$ denote the total number of users, items, and lists, respectively. Let $\U = \{u_1, u_2, ..., u_{n_U}\}$ and $\I = \{i_1, i_2, ..., i_{n_I}\}$ denote the sets of users and items, respectively. Also, let $\calL = \{\bfL^{u_1}, \bfL^{u_2}, ..., \bfL^{u_{n_U}}\}$ denote the collection of sets of users' lists, where, $\bfL^u = \{l_1^u, l_2^u, ..., l_{\eta(u)}^u\}$ denotes the $\eta(u)$ number of lists that the user $u$ has created. For each user $u \in \U$, the $j^\mathrm{th}$ list in $\bfL^u$, \textit{i.e.}, $l_j^u$, contains items ordered based on their timestamps. Given the data $\{(u_k, \bfL^{u_k})\}_{k=1}^{n_U}$, the aim of the personalized list continuation task is to find an $i \in \I$ which has a high probability of getting selected and added by the user $u$ to the list $l_j^u$, for every user-list pair $(u, l_j^u)$. To simplify our notations, we use $(u, i, l)$ to denote user, item, and list indices (IDs) throughout this section, where $l \in \bfL^u$ and item $i$ belongs to the list $l$. Also, we drop all the bias weight vectors from the equations below for ease of explanation.

\begin{figure*}[h]
    \centering
    \begin{subfigure}{0.9\textwidth} 
        \includegraphics[height=240pt]{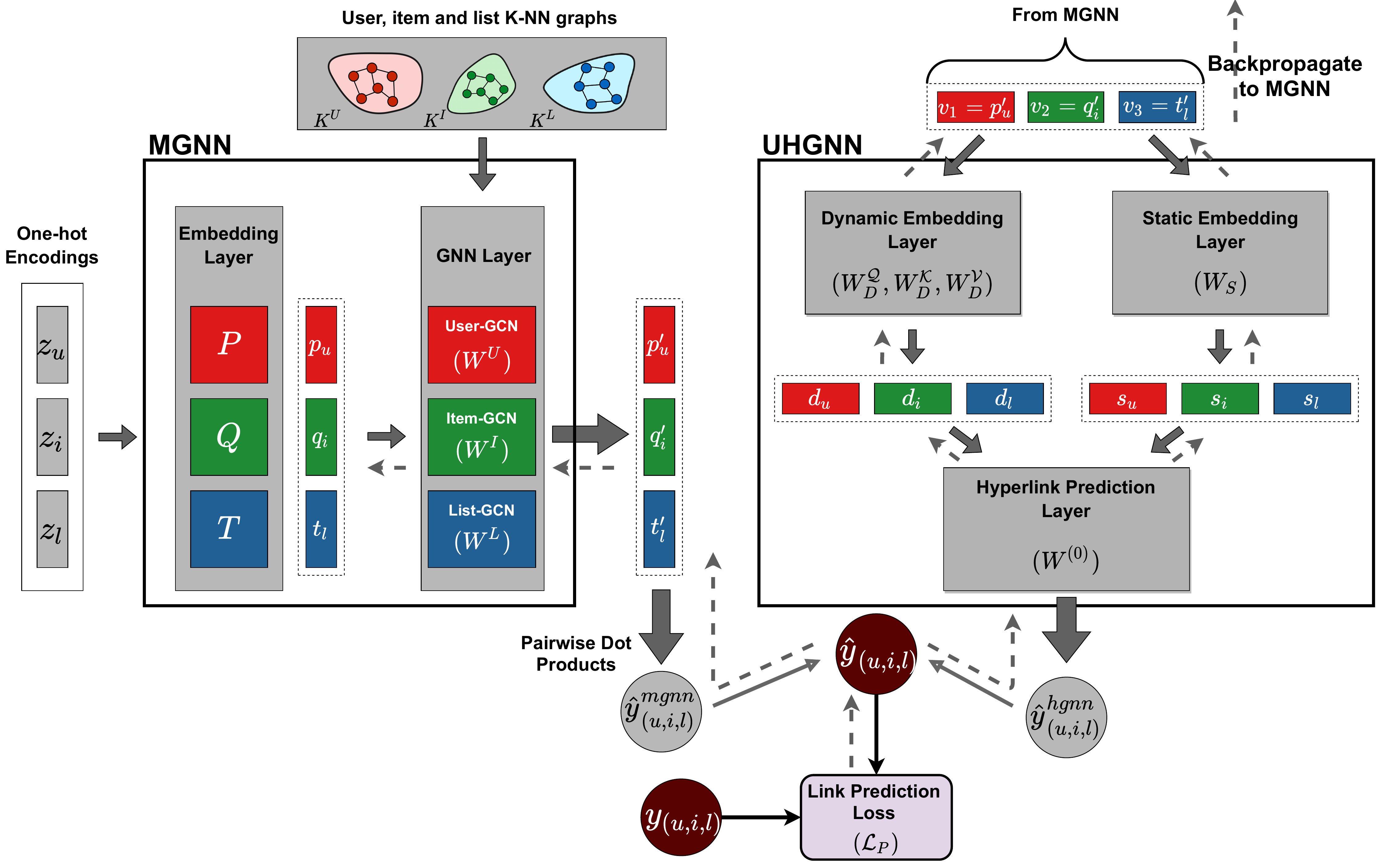}
        \caption{Multi-view Graph (MGNN) and Uniform Hypergraph (UHGNN) Neural Networks} 
        \label{fig:architecture_a}
    \end{subfigure} 
    \hfill
    \begin{subfigure}{\textwidth} 
        \includegraphics[height=190pt]{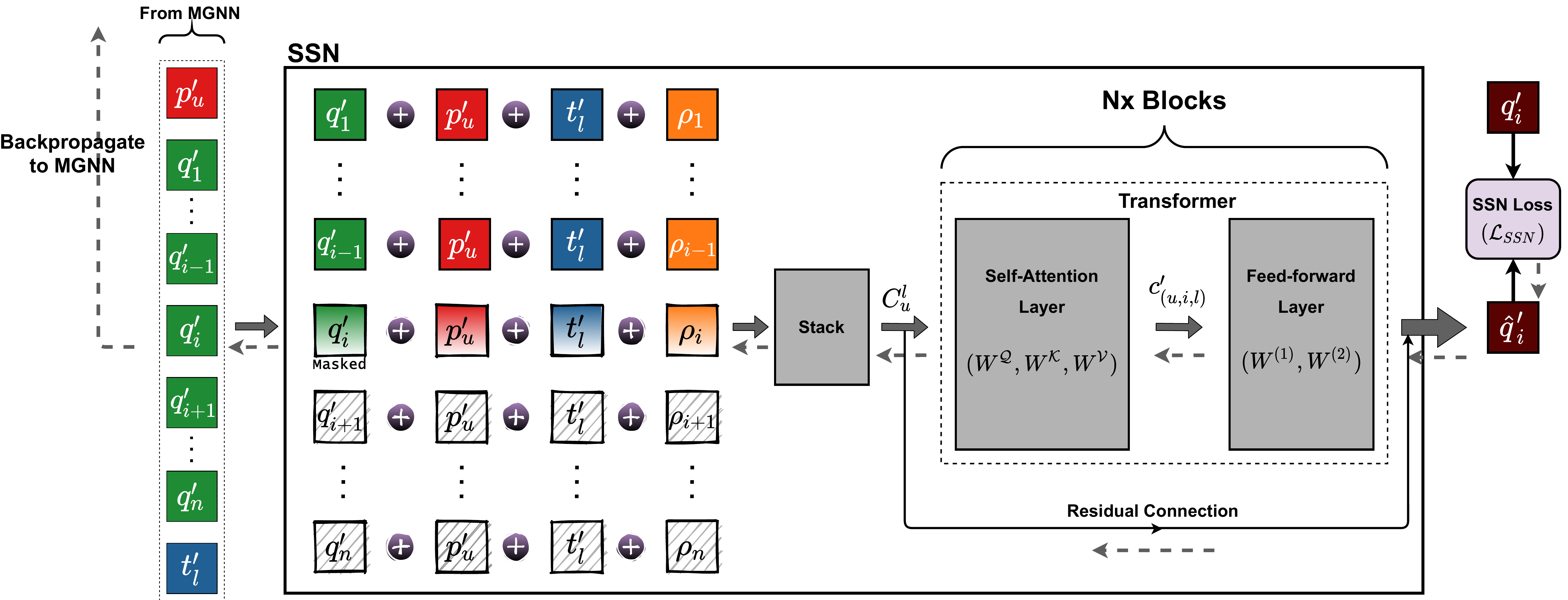}\hfill
        \caption{Self-Attention Sequence Network (SSN)}
        \label{fig:architecture_b}
    \end{subfigure} 
    \hfill
    \caption{Illustration of the HyperTeNet architecture. Networks in (a) capture homogeneous and heterogeneous relationships among entities, while (b) captures the sequential nature of lists. The training algorithm alternates between (a) and (b) to learn better representations $p_u$, $q_i$, and $t_l$. Backward dashed lines indicate the error backpropagation.}
    \label{fig:hypertenet}
\end{figure*}

Now, we explain our proposed model in detail. Our model consists of three sub-models as illustrated in \autoref{fig:hypertenet}: (a) Multi-view Graph Neural Network (MGNN), that learns relationships among the same types of entities; (b) Uniform Hypergraph Neural Network (UHGNN), that exploits ternary relationships amongst users, items, and lists; and (c) Self-attention Sequence Network (SSN) that captures sequence information from items in a list.

\subsection{Multi-view Graph Neural Network (MGNN)}
First, we randomly initialize the embeddings for users, items, and lists. Then, we construct $K$-Nearest Neighbor ($K$-NN) graphs for users, items, and lists, individually. Finally, graph convolutions are applied to individual graphs to learn relationships among the same types of entities.

\subsubsection{Initial Embedding Layer}
Let $P \in \mathbb{R}^{\embeddim \times n_U},  Q \in \mathbb{R}^{\embeddim \times n_I}$ and $T \in \mathbb{R}^{\embeddim \times n_L}$ be randomly initialized embedding matrices, with embedding dimension $\embeddim$, for users, items and lists respectively. Let $z_u \in \mathbb{R}^{n_U}, z_i \in \mathbb{R}^{n_I}, z_l \in \mathbb{R}^{n_L}$ be one-hot encoding vectors for user $u$, item $i$ and list $l$, respectively. We construct user embedding ($p_u$), item embedding ($q_i$) and list embedding ($t_l$) from $P$, $Q$ and $T$ as follows:
\begin{align}
    \begin{split}
        p_u = P z_u, ~
        q_i = Q z_i, \text{~and~}
        t_l = T z_l.
    \end{split}
\end{align}

\subsubsection{K-Nearest Neighbor Graph}
HyperTeNet creates $K$-NN graphs for users, items, and lists to learn the relationship among the same types of entities. To do this, we first start with the user-list, list-item, and user-item interaction sets and create list-item and user-item bipartite graphs from them. The nodes of the two graphs are $\calL \cup \I$, and $\U \cup \I$, respectively. An edge in the list-item (or user-item) bipartite graph connects a list (or a user) to an item. By taking a random walk on these bipartite graphs, we create a node sequence. For example, the node sequence generated by taking a random walk on the list-item graph consists of alternating list and item entities. We then apply the skip-gram technique \cite{mikolov2013distributed} for getting representations for the entities. These representations indicate how closely two entities are related. Further, using these representations, we identify the $K$ nearest neighbors of the entities and construct $K$-Nearest Neighbor ($K$-NN) graphs, with adjacency matrices $K^U$, $K^I$ and $K^L$ for users, items, and lists, respectively.

\subsubsection{Graph Neural Network Layer}
We employ a graph convolution network (GCN) to incorporate information from multi-hop neighbors of the same entity type. For this, we start with the $K$-NN graphs constructed above and apply graph convolutions on each of the graphs individually, taking initial embeddings as the input to the nodes. A single GCN layer updates the entity embeddings using the embeddings from neighbors of the same type as
\begin{align}
    \begin{split}
        p'_u &= \sum_{k \in N^{U}(u) \cup \{u\}} K^{U}_{uk} W^U \cdot p_k, \\
        q'_i &= \sum_{k \in N^{I}(i) \cup \{i\}} K^{I}_{ik} W^I \cdot q_k, \\
        t'_l &= \sum_{k \in N^{L}(l) \cup \{l\}} K^{L}_{lk} W^L \cdot t_k,
    \end{split}
    \label{eqn:mgnn_embeds}
\end{align}
where $N^U(u), N^I(i)$ and $N^L(l)$ denote the neighbors of user $u$, item $i$ and list $l$, respectively. Further, $W^U, W^I$ and $W^L$ are the weight matrices corresponding to the three GCNs, and $p'_u, q'_i$ and $t'_l$ are the updated embeddings for user $u$, item $i$ and list $l$. We use multiple GCN layers and similar procedure is followed in other layers.%Here, depth of the layers indicate incorporation of information from higher-order neighbors. 

\subsubsection{Dot Product Layer}
Apart from producing the GCN-updated embeddings, $p'_u$, $q'_i$ and $t'_l$, that act as inputs to UHGNN and SSN, this network also predicts pairwise links between user $u$ and item $i$, and list $l$ and item $i$ via the dot product layer. The dot product layer takes the pairwise inner products of the three embeddings, \textit{i.e.}, $p'_u$ and $q'_i$, $q'_i$ and $t'_l$, and $t'_l$ and $p'_u$. The resultant scalars are then combined to return a single score, $\hat{y}^{mgnn}_{(u,i,l) \in [0, 1]}$, as follows:
\begin{align}
    \label{eqn:mgnn_pred}
    \hat{y}^{mgnn}_{(u,i,l)} = \Sigmoid(p'_u \cdot q'_i + q'_i \cdot t'_l + t'_l \cdot p'_u).
\end{align}
We describe the effectiveness of $\hat{y}^{mgnn}_{(u,i,l)}$ and how it is combined with UHGNN's prediction in the next subsection. 

\subsection{Uniform Hypergraph Neural Network (UHGNN)}
For personalized list continuation, understanding the ternary relationship among the interacting entities is critical. Hence, capturing the relationship among user behaviors, item properties, and list styles via their respective embeddings becomes significant. We pose the problem as a hyperlink prediction task in a 3-uniform heterogeneous hypergraph to understand this ternary relationship. During each mini-batch in training MGNN, the graph convolutions first learn the proximity (or similarity) information from neighbors of the same (homogeneous) entity type through pairwise relationships. These are reflected in the embeddings $p'_u$, $q'_i$, and $t'_l$, which further act as inputs to UHGNN. In UHGNN, we capture the ternary relationships across different entity types (heterogeneous). Intuitively, if a hyperedge connects the three entities, it signifies that the item $i$ aligns with user $u$'s behaviors and the stylistic trend of list $l$.

In a $k$-uniform hypergraph, all the hyperedges within the hypergraph have the same size (dimension) of $k$ \cite{zhang2020hypersagnn}. A homogeneous hypergraph consists of a single type of node, while heterogeneous hypergraphs consist of two or more types of nodes. In our case, the hypergraph nodes are the users, items, and lists, making the hypergraph heterogeneous. The GCN-modified entity embeddings,  $p'_u$, $q'_i$ and $t'_l$, represent the node features. Moreover, each candidate hyperedge spans a user node, an item node, and a list node, making the hypergraph 3-uniform.

Formally, we define our 3-uniform hypergraph as $G = (V, E)$, where $V = \mathcal{U} \cup \mathcal{I} \cup \mathcal{L}$ denotes the set of nodes in the graph, and $E = \{\mathbf{e} = (u,i,l) : u \in \U, l \in \bfL^u, i \in l\}$ represents the set of hyperedges. Each hyperedge in $E$ spans exactly three nodes, \textit{i.e.}, $\delta(\mathbf{e}) = 3, \forall ~\mathbf{e} \in E$.

UHGNN is inspired by the Hyper-SAGNN architecture \cite{zhang2020hypersagnn}. Unlike hypergraph algorithms that use clique expansion \cite{hgnn_spectral_learning, hgnn_cluster_classify, hgnn_DHNE}, Hyper-SAGNN attends over a fixed set of nodes to predict whether a hyperedge connects them. While doing so, it treats the nodes symmetrically. We learn two functions that take the GCN-modified node embeddings, $p'_u$, $q'_i$, and $t'_l$, as inputs and collectively return the probability of these nodes forming a hyperedge. For ease of notation, let $v_1 = p'_u$, $v_2 = q'_i$, and $v_3 = t'_l$ denote the embedding vectors of the three nodes in the ternary interaction. The input embedding triple, $(p'_u, q'_i, t'_l)$, is passed through two parallel networks to get two resultant embeddings: static and dynamic.

\subsubsection{Static Embedding Layer}
For getting the static embeddings, the triple is fed to a position-wise feed-forward network to produce $(s_1, s_2, s_3)$, given by \autoref{eqn:hgnn_static_embeds}. Here, $s_j$ is referred to as the static embedding for node $j$ as irrespective of the input interaction triple, $s_j$ does not change for node $j$. The static embeddings, hence, express the global behavior of the nodes and are obtained as follows:
\begin{align}
    \begin{split}
        s_j = \mathrm{tanh}(W_S^T v_j), \\ 
	    \text{~~where~~} 1 \leq j \leq 3.
	\end{split}
    \label{eqn:hgnn_static_embeds}
\end{align}

\subsubsection{Dynamic Embedding Layer}
Simultaneously, the triple $(p'_u, q'_i, t'_l)$ is also given as input to the multi-head attention layer to generate the dynamic embeddings, $(d_1, d_2, d_3)$. The dynamic embeddings are termed so because they are dependent on the current node embeddings of the input triple, and hence, vary with each interaction. They capture the local behavior of nodes within an interaction. We describe the self-attention mechanism \cite{vaswani2017attention} on our hypergraph, using terms identical to the original work. Given a group of node embeddings $(v_1, v_2, v_3)$ and trainable weight matrices $W^{\mathcal{Q}}_D$, $W^{\mathcal{K}}_D$, $W^{\mathcal{V}}_D$ representing linear transformations of features, we first compute the normalized attention coefficients, $\alpha_{jk}$, as follows: 

\begin{align}
	\begin{split}
	        e_{jk} &= (W^{\mathcal{Q}}_D \cdot v_j) \cdot (W^{\mathcal{K}}_D \cdot v_k), 1 \leq j,k \leq 3, \text{~~and~~} \\
        	\alpha_{jk} &= \frac{exp(e_{jk})}{\sum_{1 \leq k' \leq 3} exp(e_{jk'})},
	\end{split}
\end{align}
where $e_{jk}$ denotes the unnormalized energy coefficient corresponding to indices $j$ and $k$. The weighted sum of the transformed features with an appropriate activation gives us the dynamic embedding for node $j$ in the interaction. 
\begin{align}
	d_j = \mathrm{tanh}\left(\sum_{1 \leq k \leq 3, j \neq k} \alpha_{jk} W^{\mathcal{V}}_D \cdot v_k \right).
\end{align}
We use a similar motivation as in Hyper-SAGNN and CBOW \cite{zhang2020hypersagnn, mikolov2013efficient} to not include $\alpha_{jj}$ in the calculation of $d_j$, \textit{i.e.}, given only the context nodes in the interaction, predict the dynamic embedding of the target node.

\subsubsection{Hyperlink Prediction Layer}
Using the static and dynamic embeddings for each node in the interaction, we calculate the element-wise power of the difference of the corresponding static-dynamic pair. It is then further passed through a one-layered neural network having sigmoid activation with weight matrix $W^{(0)}$ to produce a probability score $o_j \in [0, 1]$, corresponding to each node. We finally take the simple average of the three probabilities to get the probability of the existence of a hyperedge.
\begin{align}
    \label{eqn:hgnn_pred}
    \begin{split}
        o_j &= \Sigmoid\left(W^{(0)} \cdot (d_j - s_j)^{\circ2}\right), \text{~and}\\
	    &\hat{y}^{hgnn}_{(u,i,l)} = \frac{1}{3}\sum_{j=1}^{3}o_j,
    \end{split}
\end{align}
where $(\cdot)^{\circ2}$ denotes taking element-wise squares, and $W^{(0)}$ is the weight matrix associated with this layer.

\vspace{0.3cm}
\noindent
\textbf{Loss function.}
We take the mean of the two predicted scores, given in Equations \ref{eqn:mgnn_pred} and \ref{eqn:hgnn_pred}, to get the final score, $\hat{y}_{(u,i,l)} \in [0,1]$, for the link prediction task,
\begin{align}
    \hat{y}_{(u,i,l)} = \frac{1}{2} \left(\hat{y}^{mgnn}_{(u,i,l)} + \hat{y}^{hgnn}_{(u,i,l)}\right).
\end{align}

$\hat{y}^{mgnn}_{(u,i,l)}$ captures the pairwise relationships among users, items, and lists. Along with these pairwise relationships, we also capture the ternary relationships among the interacting entities via $\hat{y}^{mgnn}_{(u,i,l)}$. Empirically, we found that using the pairwise relationships or the ternary relationships alone do not lead to optimal results, and rather their combination betters the state-of-the-art models.

We employ the cross entropy loss for learning parameters of MGNN and UHGNN. Let $\D^{+}$ be the set of all positive user-list-item interactions of the form $(u, i, l)$, such that for user $u$ and one of her lists $l$, $i \in l$. We adopt the negative sampling strategy \cite{mikolov2013distributed} during training, with $\D_{G}^{-}$ denoting the set of negatively sampled interactions of the form $(u, i', l)$, such that $i' \not\in l$. Section \ref{section:exp_settings} discusses the optimal cardinality of $\D_{G}^{-}$. The training set for MGNN and UHGNN is, thus, $\D_{G} = \D^{+} \cup \D_{G}^{-}$. Let $\IndicatorUIL \in \{0, 1\}$ be the indicator that an interaction $(u, i, l) \in \D^{+}$. The loss function for the two networks is then given as

\begin{align}
    % \begin{split}
        \min_{\mathcal{W}_{G}} ~ \mathcal{L}_{P}(\mathcal{W}_{G}) = &-\sum_{(u,i,l) \in {\D_{G}}} \biggl(\IndicatorUIL \ln (\hat{y}_{(u,i,l)}) \nonumber\\ &+
        \IndicatorUILGNeg \ln (1 - \hat{y}_{(u,i,l)})\biggr)
     \label{eqn:hgnn_loss}
    % \end{split}
\end{align}
where $\mathcal{W}_{G}$ consists of all the model parameters corresponding to MGNN and UHGNN.

\subsection{Self-Attention Sequence Network (SSN)}
\autoref{fig:architecture_b} illustrates the self-attention sequence component of the HyperTeNet model, abbreviated as SSN. The objectives of this network are two-fold: (a) to identify the influential items responsible for defining a list using the attention mechanism, and (b) to learn the sequential information present among the items in the list. In the setting described earlier, each user has a collection of lists, and each list consists of an ordered sequence of items.

Consider a user $u \in \U$ who has created an item list $l \in \bfL^u$. We consider the items in $l$ to be sorted in the chronological order of the timestamps in which they were added to $l$. The embeddings for $u$ and $l$ are $p_u$ and $t_l$, respectively. Let us assume that $(1, 2, ..., n)$ are the indices of the sequence of items in the ordered list $l$, with the item embeddings being $q_1, q_2, ..., q_n$ respectively. We feed all of these embeddings to MGNN's Graph Neural Network layer to extract the updated embeddings, $p'_u$, $t'_l$, and $q'_1, q'_2, ..., q'_n$. This is a modeling choice that gives better results than the plain embeddings and is shown in \autoref{fig:modeling_choice}. The updated embeddings are, thus, given as $p'_u$ for $u$, $t'_l$ for $l$ and $q'_1, q'_2, ..., q'_n$ for the $n$ items in $l$. 

Next, for each item $i \in l$, we perform $c_{(u, i, l)} = q'_i + p'_u + t'_l$. Intuitively, this operation combines the effect of user $u$'s behavior, item $i$'s properties, and list $l$'s style to give us a single vector associated with each item in the list. However, the sequential nature of $l$ is still not captured in $c_{(u, i, l)}$. In order to encode the sequential information, SSN infuses the learnable positional embeddings $\rho_1, \rho_2, ...,\rho_n$ into the corresponding $c_{(u, ., l)}$ vectors, where $\rho_t \in \R^{\embeddim}$. We can stack up the \textit{combined representations} with positional information for each $i \in l$ in a matrix, given as
\begin{align}
    \begin{split}
        C_l^u = \begin{bmatrix}
                    q'_1 + p'_u + t'_l + \rho_1 \\
                    q'_2 + p'_u + t'_l + \rho_2 \\
                    \cdot \cdot \cdot \\
                    q'_n + p'_u + t'_l + \rho_n \\
                \end{bmatrix}
    \end{split}
\end{align}
The combined input representation matrix, $C_l^u$, contains the net effect of different embeddings. It is passed through multiple self-attention layers and feed-forward neural network layers as illustrated in \autoref{fig:hypertenet}. $C_l^u$, thus, acts as inputs to the subsequent layers.

\subsubsection{Self-Attention Layer}
Similar to the maskings in BERT \cite{devlin2018bert}, we mask the last $(n-i)$ combined embeddings at the input, for $i \in \{2, ..., n\}$, and predict the item embedding $\hat{q}'_i$ at the output. The input embeddings are fed in an autoregressive manner, \textit{i.e.}, we use the items present before the masked combined embedding to predict the masked item. $C_l^u$ is first passed through a self-attention network $\SA( \cdot)$. Let $W^\mathcal{Q}, W^\mathcal{K}$, $W^\mathcal{V} \in \R^{\embeddim \times \embeddim}$ be the weight matrices. Then, we obtain the output representation $C_l^{'u}$ as follows:
\begin{align}
    \begin{split}
        C_l^{'u} = \SA(C_l^u) = \Attention(C_l^u \cdot W^\mathcal{Q}, C_l^u \cdot W^\mathcal{K}, C_l^u \cdot W^\mathcal{V}), \\
        \text{where~~}~~ \Attention(\mathcal{Q},\mathcal{K},\mathcal{V}) = \Softmax\left(\frac{\mathcal{Q} \cdot \mathcal{K}^\texttt{T}}{\sqrt{d}}\right) \cdot \mathcal{V}.
    \end{split}
\end{align}
Note that the autoregressive mask forbids the information leak between $\mathcal{Q}_i$ and $\mathcal{K}_j$, for $j > i$.

\subsubsection{Feed-forward Layer}
Let $c'_{(u,i,l)}$ be the $i^{\mathrm{th}}$ row vector of $C_l^{'u}$, corresponding to the masked item $i$. We apply a feed-forward network to $c'_{(u,i,l)}$, with shared parameters as follows:
\begin{align}
    \begin{split}
        \hat{q}'_i &= \FF(c'_{(u,i,l)}) \\ 
        &= \ReLU\left(W^{(1)}c'_{(u,i,l)}\right) W^{(2)},
    \end{split}
\end{align}
where $W^{(1)}$ and $W^{(2)}$ are weight matrices shared across all the rows of $C_l^{'u}$. Note that $\hat{q}'_i$ is a function of $u$, $i$, and $l$, emphasizing that the predicted vector accommodates personalization. The Self-Attention and Feed-forward layers together make up a Transformer block, as illustrated in \autoref{fig:architecture_b}. There can be multiple such Transformer blocks included in SSN, and we set that as a hyperparameter (discussed in Section \ref{section:exp_settings}).

\subsubsection{Regularization}
We employ dropout, residual connections, and layer-norm to alleviate overfitting, instability in training due to vanishing gradients, and higher training time. Let $g(x)$ denote the self-attention layer or the feed-forward layer. We regularize each of the two networks to get $g'(x)$,
\begin{align}
    \begin{split}
        g'(x) = x + \Dropout(g(\Layernorm(x))),
    \end{split}
\end{align}
where $g'(x)$ is the resultant output representation. Finally, the masked items (combined representations) at the inputs are predicted at the output, and during inference, these serve as potential candidate recommendations for getting added to their respective lists.

\subsubsection{Relevance Prediction Layer}
Let $r_i$ denote the relevance score of item $i$ being the next item in list $l$ for user $u$, given the first $i-1$ items of $l$. To avoid notational clutter, we drop $l$ and $u$ from the relevance score. Also, recall that $q'_i$ is the GCN-updated embedding for item $i$, obtained in \autoref{eqn:mgnn_embeds}. To find the relevance score $r_i$, we take the sigmoid of inner product of the output from the final feed-forward layer, $\hat{q}'_i$, and $q'_i$:
\begin{align}
    r_i = \Sigmoid \left(q'_i \cdot \hat{q}'_i \right).
    \label{eqn:ssn_relevance_score}
\end{align}
A high interaction score, $r_i$, means high relevance of item $i$ being the next item in the list, and we can generate recommendations by ranking these scores.

\vspace{0.3cm}
\noindent
\textbf{Loss function:} Like in the loss function defined for the two graph neural networks, let $\D^{+}$ and $\D_{SSN}^{-}$ be the set of all positive interactions and the set of negatively sampled interactions, respectively. It is important to note that the negative samples (and even the number of negative samples) in $\D_{G}^{-}$ and $\D_{SSN}^{-}$ need not be the same. We treat this number as a hyperparameter and discuss it in Section \ref{section:exp_settings}. The training set for SSN is $\D_{SSN} = \D^{+} \cup \D_{SSN}^{-}$, with $\IndicatorUIL \in \{0, 1\}$ being the indicator that an interaction $(u, i, l) \in \D^{+}$. We define the binary cross entropy-based criterion for this network as follows: 
\begin{align}
    % \begin{split}
        \min_{\mathcal{W}_{SSN}} ~ \mathcal{L}_{SSN}(\mathcal{W}_{SSN}) = &-\sum_{(u,i,l) \in {\D_{SSN}}} \biggl(\IndicatorUIL \ln (r_i) \nonumber\\ &+ \IndicatorUILSSNNeg \ln(1 - r_i)\biggr)
    \label{eqn:ssn_loss}
    % \end{split}
\end{align}
where $\mathcal{W}_{SSN}$ consists of all the model parameters associated with SSN. 

\vspace{0.3cm}
\noindent
\textbf{Inference:} During inference, for an input user $u$ and list $l$, such that $l \in \bfL^u$, we first use MGNN to get the representations $p'_u$, $t'_l$, and $q'_1, ..., q'_n$, where the list $l$ has $n$ items. Also, UHGNN is inactive and we do not predict $\hat{y}^{(u,i,l)}$. The embedding vectors are then fed to SSN as inputs, which predicts the $(n+1)^{\mathrm{th}}$ item embedding vector, $\hat{q}'_{(n+1)}$. Next, we calculate the reference scores for each item in the candidate set of items of $u$ and $l$, given by \autoref{eqn:ssn_relevance_score}. Lastly, we rank the candidate items based on the obtained relevance scores.

We now show the effectiveness of HyperTeNet by comparing it with some state-of-the-art baseline models. Through various ablation studies, we also demonstrate the potency of each of HyperTeNet's sub-models and our modeling choices.

\section{Experimental Setup}
This section discusses the datasets we use; the evaluation protocol we follow; the baseline models against which we gauge HyperTeNet's performance; and the hyperparameter settings for reproducibility.

\subsection{Datasets Used}
    We perform our experiments on four real-world datasets, covering a wide range of user-generated lists (book-based, song-based, and answer-based). Since all of these crawled datasets have a large number of user-list-item interactions, we have created one random subset from each of them to save computational costs and demonstrate the approach. 
        \subsubsection{\textbf{Art of the Mix}}
        Art of the Mix\footnote{http://www.artofthemix.org/} (AOTM) is a website that contains music playlists, populated by its users since 1997. McFee and Lanckriet\footnote{https://brianmcfee.net/data/aotm2011.html} \cite{mcfee2012hypergraph} scraped the website since its inception till 2011 for playlists.
        \subsubsection{\textbf{Goodreads}}
        Goodreads is a popular site to rate books while also allowing users to create book-based lists.
        \subsubsection{\textbf{Spotify}}
        Spotify is an audio streaming provider that enables its users to create, edit, and share song-based playlists.
        \subsubsection{\textbf{Zhihu}}
        Zhihu is a question-and-answer website where users can ask and answer questions to the community. The dataset contains correlated answers (connected by tags/common spaces) in the form of lists.

    \autoref{tab:dataset_statistics} summarizes the statistics of the obtained subsets. Apart from AOTM, all the other datasets have been scraped by He et al.\footnote{https://github.com/heyunh2015/ListContinuation\_WSDM2020} \cite{he2020consistency}. To compare HyperTeNet with the baseline models available in the literature, we run each model five times on the created subset and take the average of the scores obtained. The following subsection discusses the evaluation protocol to obtain those scores.

    \begin{table}[h]
    \begin{center}
    \caption{Dataset Statistics}
    \label{tab:dataset_statistics}
    \resizebox{\columnwidth}{!}{
        \begin{tabular}{c|c|c|c|c}
            \hline
            \multirow{2}{*}{Statistic} & \multirow{2}{*}{~AOTM~} & \multirow{2}{*}{~Goodreads~} & \multirow{2}{*}{Spotify} & \multirow{2}{*}{~Zhihu~} \\
            & & & & \\
            \hline
            %\multirow{4}{*}{Amazon}
            %------
            ~~\#Users~~                     & 1,134     & 754       & 347     & 1,104 \\
            ~~\#Lists~~                     & 16,613    & 2,881     & 4,165   & 8,997 \\
            ~~\#Items~~                     & 15,759    & 6,733     & 31,958  & 10,937 \\
            ~~\#List-Item interactions~~    & 169,885   & 178,178   & 318,595 & 178,208 \\
            ~~Avg. \#Lists per user~~       & 14.66     & 3.82      & 12.03   & 8.16 \\
            ~~Avg. \#Items per list~~       & 10.23     & 61.87     & 76.51   & 19.81 \\
            ~~Density~~                     & 0.065\%   & 0.919\%   & 0.239\% & 0.181\% \\
             \hline\hline
        \end{tabular}
    }
    \end{center}

\end{table}

    \subsection{Evaluation Criteria}
    On very large-scale real-world datasets, it is infeasible to rank the whole millions of items for final recommendations. Hence, generating recommendations follows generating a much smaller candidate set from the dataset and ranking items within the subset. The baselines and this work, however, do not deal with candidate set generation. Following \cite{he2020consistency,tran2019adversarial,he2017neural}, we use the \textit{leave-one-out} strategy to evaluate the performance of the proposed model against the comparison models. First, lists are represented based on the order in which items are added to the lists. Then, we remove the last item and previous last item from each list and keep them in the test and validation set, respectively. The remaining items in the lists are used for training the model. Since it is time-consuming to rank the items for evaluation against all the available items in the system, we generate a candidate set by randomly sampling 100 negative items for every ground-truth item and ranking them based on their predicted scores. Consequently, the validation and test sets have 101 items for a user-list pair, containing the ground-truth item and 100 negative items.
    
    We employ three widely used top-N recommendation metrics for performance comparison \cite{he2020consistency, tran2019adversarial, he2017neural}. Specifically, we select $N=5$ ranked items from the set of candidate items for each list and compare them against the ground-truth item. We utilize three evaluation metrics: Hit Ratio at 5 (HR@5), Normalized Discounted Cumulative Gain at 5 (NDCG@5), and Mean Average Precision at 5 (MAP@5). Here, in the recommended list with $N=5$ items, HR@N measures the presence of the positive item, MAP@N computes the mean of the Average Precision (AP) over all the users, and NDCG@N accounts for the position of the positive item.

    \subsection{Comparison Models} 
    We compare HyperTeNet with the following models:
        \subsubsection*{\textbf{CAR~\cite{he2020consistency}}} Consistency-aware and Attention-based Recommender (CAR) utilizes an attention and gating mechanism to learn the consistency strength of the lists. Further, the consistency strength is attentively exploited for item list continuation.
        
        \subsubsection*{\textbf{AMASR~\cite{tran2019adversarial}}} The Adversarial Mahalanobis distance-based Attentive Song Recommender (AMASR) tackles the personalized playlist continuation task by leveraging memory attention mechanisms and the Mahalanobis distance-based metric learning approach to learn user preferences and song playlist themes. Tran et al. \cite{tran2019adversarial} also propose two other models: AMDR and AMASS, which we include in our comparisons.
        
        \subsubsection*{\textbf{SASRec~\cite{kang2018self}}} Self-Attention based Sequential Recommendation (SASRec) uses attention mechanisms to capture long-term sequences by attentively selecting few actions for next item recommendation.
        
        \subsubsection*{\textbf{Caser~\cite{tang2018personalized}}} Convolutional Sequence Embedding Recommendation Model (Caser) is proposed for sequential recommendation. It embeds a sequence of items into an image and learns the user's general preferences and sequential patterns using horizontal and vertical convolutional filters. 

        \subsubsection*{\textbf{NeuMF and GMF~\cite{he2017neural}}} Neural Matrix Factorization (NeuMF) is based on a wide and deep framework proposed for implicit rating setting with only user-item interactions. It combines the representations learned from both matrix factorization and deep neural networks to get rich representations of user-item interactions. Generalized Matrix Factorization (GMF) is a sub-model of NeuMF, which, as the name suggests, generalizes Matrix Factorization (MF) via a single-layer neural network.
        
        \subsubsection*{\textbf{BPR~\cite{rendle2009bpr}}} Bayesian Personalized Ranking (BPR) is a standard baseline for top-N recommendation setting. It employs a pairwise loss function for ranking settings during optimization.

    \begin{table*}
	\caption{Performance of various models for personalized list continuation task on four real-world datasets -- AOTM, Goodreads, Spotify, and Zhihu. Boldfaced values indicate the best model along a dataset and a metric (\textbf{HR@5}, \textbf{MAP@5} and \textbf{NDCG@5}). ``Improvement" indicates the percentage increase in performance over the next best model (denoted by asterisk). We conduct paired \textit{t}-test, and the improvements using HyperTeNet are statistically significant with $p < 0.01$. ``Std. dev." signifies the standard deviation of HyperTeNet's results. We omit the standard deviations of other models due to space constraints.}
	\label{tab:performance}
	\resizebox{\textwidth}{!}{
		\begin{tabular}{c||ccc|ccc|ccc|ccc}
			\toprule
			\multirow{2}{*}[-1pt]{Model} & \multicolumn{3}{c|}{AOTM} & \multicolumn{3}{c|}{Goodreads} & \multicolumn{3}{c|}{Spotify} & \multicolumn{3}{c}{Zhihu} \\
			& HR@5 & NDCG@5 & MAP@5 & HR@5 & NDCG@5 & MAP@5 & HR@5 & NDCG@5 & MAP@5 & HR@5 & NDCG@5 & MAP@5 \\
			\midrule \midrule
			BPR~\cite{rendle2009bpr} & 0.2506 & 0.1783 & 0.1546 & 0.5193 & 0.3724 & 0.3230 & 0.5017 & 0.3626 & 0.3070 & 0.5309 & 0.3964 & 0.3520 \\
			GMF~\cite{he2017neural} & 0.2669 & 0.1794 & 0.1507 & 0.4867 & 0.3398 & 0.2904 & 0.4711 & 0.3219 & 0.2694 & 0.4780 & 0.3404 & 0.2949 \\
			NeuMF~\cite{he2017neural} & 0.2689 & 0.1801 & 0.1513 & 0.4977 & 0.3533 & 0.3053 & 0.4790 & 0.3264 & 0.2786 & 0.4829 & 0.3551 & 0.3129 \\
			Caser~\cite{tang2018personalized} & 0.2508 & 0.1745 & 0.1495 & 0.5286 & 0.3892 & 0.3410 & 0.5209 & 0.3842 & 0.3399 & 0.5174 & 0.3908 & 0.3489 \\
			SASRec~\cite{kang2018self} & 0.3515 & 0.2455 & 0.2107 & 0.5441 & 0.3945 & 0.3487 & 0.5386 & 0.3991 & 0.3505 & 0.5868* & 0.4407 & 0.3923 \\
			AMDR~\cite{tran2019adversarial} & 0.4250 & 0.3111 & 0.2735 & 0.5653 & 0.4111 & 0.3672 & 0.5267 & 0.3833 & 0.3375 & 0.5517 & 0.4203 & 0.3761 \\
			AMASS~\cite{tran2019adversarial} & 0.3703 & 0.2612 & 0.2253 & 0.4562 & 0.3124 & 0.2766 & 0.4809 & 0.3312 & 0.2838 & 0.4459 & 0.3247 & 0.2846 \\
			AMASR~\cite{tran2019adversarial} & 0.4294 & 0.3124 & 0.2737 & 0.5394 & 0.3907 & 0.3473 & 0.5192 & 0.3732 & 0.3246 & 0.5391 & 0.4070 & 0.3632 \\
			CAR~\cite{he2020consistency} & 0.4314* & 0.3135* & 0.2745* & 0.5964* & 0.4436* & 0.3921* & 0.5823* & 0.4303* & 0.3801* & 0.5816 & 0.4438* & 0.3981* \\
			\midrule
			\textbf{HyperTeNet (Ours)} & \textbf{0.4552} & \textbf{0.3297} & \textbf{0.2811} & \textbf{0.6592} & \textbf{0.4942} & \textbf{0.4394} & \textbf{0.6306} & \textbf{0.4750} & \textbf{0.4235} & \textbf{0.6204} & \textbf{0.4767} & \textbf{0.4291} \\
			\midrule
			Std. dev. & 0.0008 & 0.0006 & 0.0008 & 0.0016 & 0.0011 & 0.0012 & 0.0015 & 0.0017 & 0.0017 & 0.0016 & 0.0012 & 0.0013 \\
			\midrule
			Improvement (in \%) & 5.52 & 5.17 & 2.40 & 10.53 & 11.41 & 12.06 & 8.29 & 10.39 & 11.42 & 5.73 & 7.41 & 7.79 \\
			\bottomrule
		\end{tabular}
	}
\end{table*}

    \subsection{Experimental Settings}
    \label{section:exp_settings}
        \subsubsection*{\textbf{General Settings}}
        All the hyperparameters of HyperTeNet and baseline methods are tuned on the validation set. The training terminates when the model trains for the maximum number of epochs (set to 300) or when the validation NDCG does not improve over 20 epochs (early stopping strategy). To train HyperTeNet, we use the Adam optimizer with learning rates of 0.001, 0.008, 0.008, and 0.002 for the AOTM, Goodreads, Spotify, and Zhihu datasets, respectively. We select a batch size of 2048 for MGNN and the hypergraph network and 256 for SSN based on the validation results. Also, the maximum sequence length for all datasets is set to 300. For all the datasets, we observe that the optimal number of negative samples for every positive instance during training lies between 3 and 7. All our experiments are conducted with a single Tesla P100-PCIe-16GB GPU using the PyTorch framework. For obtaining results on AOTM, Goodreads, Spotify, and Zhihu, HyperTeNet takes 13, 13, 18, and 19 seconds per epoch, respectively.

    	\subsubsection*{\textbf{Embedding Sizes}}
    	We tried a range of values for the user, item, and list embedding sizes, starting from a lower value of 8 to an upper limit of 256. Embedding size in the range of 64-96 gave us the optimal validation performance on all four datasets.

    	\subsubsection*{\textbf{Number of Transformer Blocks}}
    	Empirically, the number of Transformer blocks in SSN is set to 2 for AOTM and Zhihu and 3 for Goodreads and Spotify. Including more than 3 blocks does not improve HyperTeNet's performance and rather increases the training duration.
    	
        \subsubsection*{\textbf{Nearest Neighbors Graph}}
    	We observed that the number of nearest neighbors, $K$, in the $K$-NN graph considerably impacts the model's performance (\autoref{fig:hyperparameter_study_a}). The peak performance for AOTM, Spotify and Zhihu is achieved at $k = 50$, while for Goodreads, the peak occurs at $k = 25$.
    	
    	\subsubsection*{\textbf{Hypergraph Neural Network}}
    	Based on the validation set performance, we set the number of heads in the multi-head attention layer to 8 and the size of hypergraph representation vectors to 64.

\section{Results and Discussion}
We now discuss the experimental results concerning addressing the following research questions:
\begin{description}
    \item{\textbf{RQ1:}} Does our proposed model perform better than the state-of-the-art models for the personalized list continuation problem?% (Section \ref{subsec:rq1})
    \item{\textbf{RQ2:}} Does an ablation study on HyperTeNet verify the effectiveness of its sub-models?
    \item{\textbf{RQ3:}} What is the effect of different hyperparameters on the performance of the model?% (Section \ref{subsec:rq3})
\end{description}

    \subsection{RQ1: Personalized List Continuation Results}
    \autoref{tab:performance} demonstrates how well HyperTeNet performs against the baseline models for the personalized list continuation task. Our experiments reveal that CAR \cite{he2020consistency} is the best baseline on all the datasets. HyperTeNet edges over CAR and considerably improves all the evaluation scores (HR, MAP and NDCG) on the test sets of all the datasets, further attesting to the ideas behind its architecture. Higher metric scores indicate that the quality of item recommendations to lists produced by HyperTeNet is very high. We conduct paired \textit{t}-test and note that the improvements using HyperTeNet are statistically significant with $p < 0.01$.
    
    The higher evaluation scores of HyperTeNet can be attributed to the fact that its learned embeddings capture the multi-hop relationships among the same-type entities and ternary relationships among users, items, and lists, while also being aware of the sequence of items in the lists. HyperTeNet's sub-models encapsulate these three ingredients, and as the results imply, these are vital for personalized list continuation.
    
    \subsection{RQ2: Ablation Study and Modeling Choices}
    % \begin{figure}[htp]
    %     \centering
    %     \begin{subfigure}{0.9\columnwidth}
    %         \includegraphics[width=\columnwidth]{figures/experiment_ablation.pdf} \hfill
    %         \caption{Network Ablation Study: each column is an overlap of evaluation scores of (i) SSN only, (ii) MGNN + SSN, and (iii) MGNN + UHGNN + SSN (HyperTeNet) networks.}
    %         \label{fig:ablation_study_a}
    %     \end{subfigure} 
    %     \hfill
    %     \begin{subfigure}{0.9\columnwidth} 
    %         \includegraphics[width=\columnwidth]{figures/experiment_embeds.pdf}\hfill
    %         \caption{Effect of using MGNN embeddings at UHGNN and SSN instead of plain embeddings.} 
    %         \label{fig:ablation_study_b}
    %     \end{subfigure} 
    %     \label{fig:ablation_study}
    %     \caption{Ablation Study and Modeling Choices}
    % \end{figure}
    
    \begin{figure}[htp]
        \centering
        \includegraphics[width=\columnwidth]{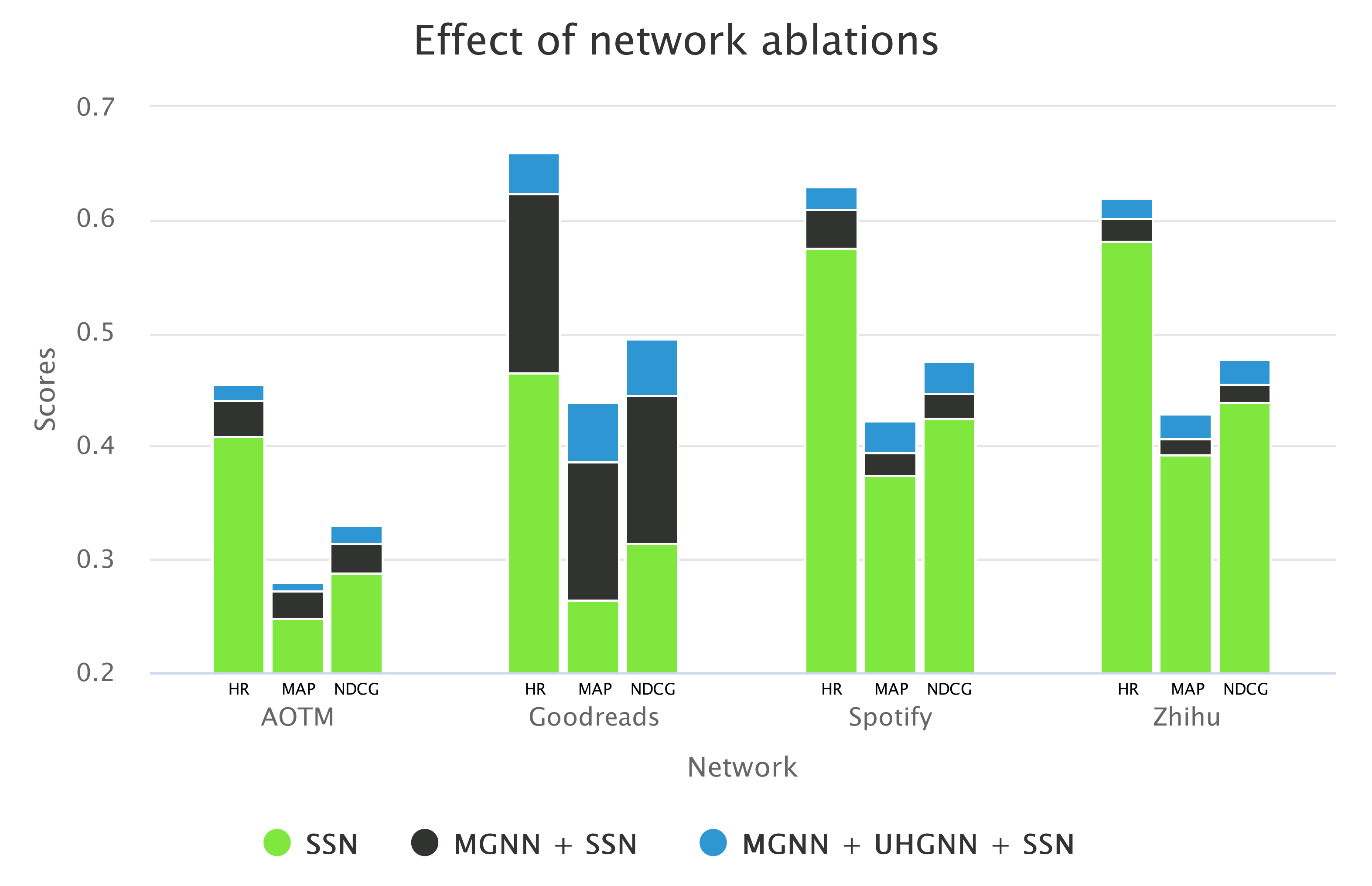}
        \caption{Network Ablation Study: each column is an overlap of evaluation scores of (i) SSN only, (ii) MGNN + SSN, and (iii) MGNN + UHGNN + SSN (HyperTeNet) networks.}
        \label{fig:ablation_study}
    \end{figure}
    
    \begin{figure}[htp]
        \centering
        \includegraphics[width=\columnwidth]{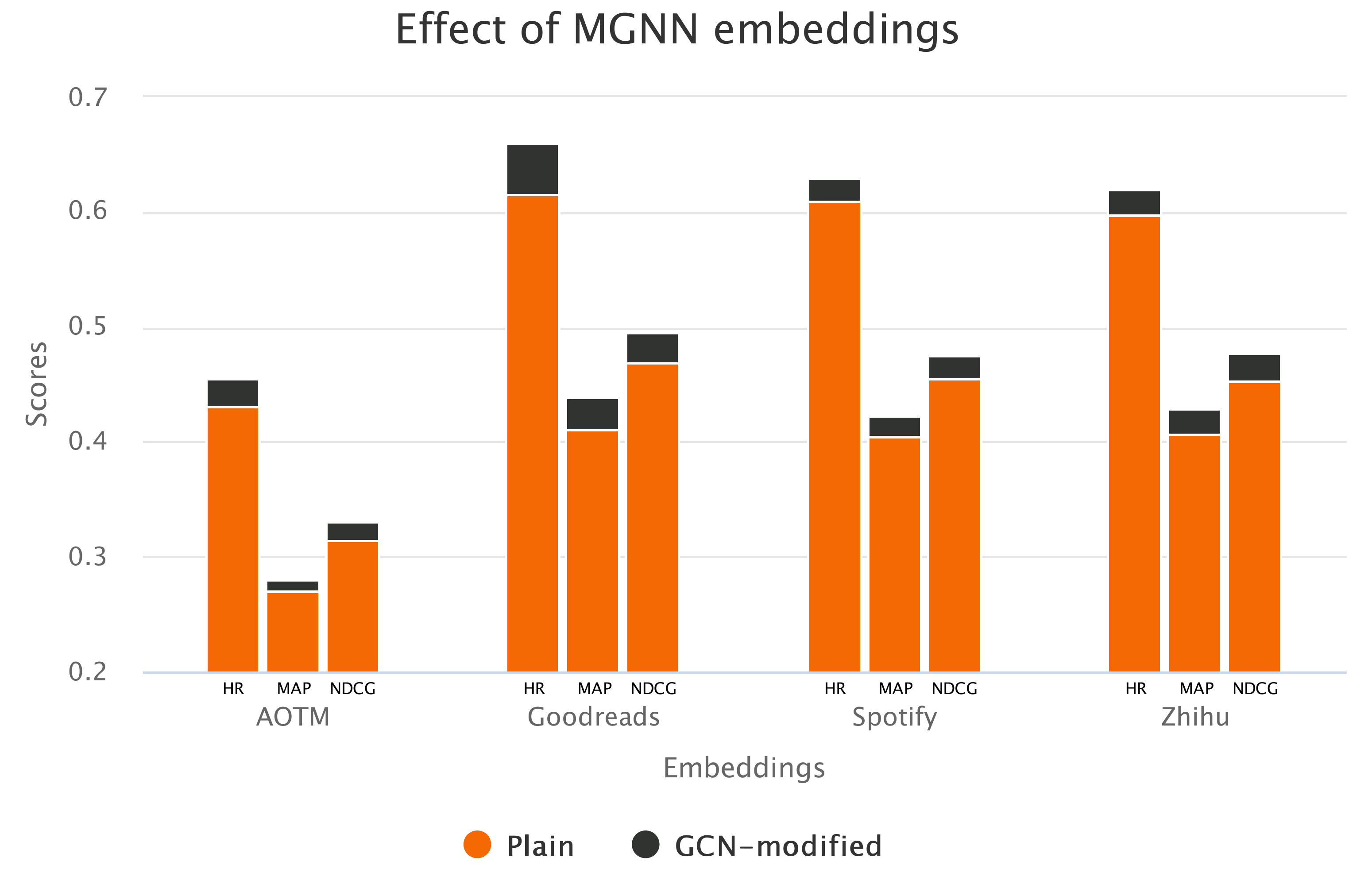}
        \caption{Effect of using MGNN embeddings $(p'_u, q'_i, t'_l)$ at UHGNN and SSN instead of plain embeddings $(p_u, q_i, t_l)$.} 
        \label{fig:modeling_choice}
    \end{figure}
    
    \autoref{fig:ablation_study} shows the ablation study results on HyperTeNet. A column indicates the observed metric scores by the three networks on a dataset.
        \subsubsection*{\textbf{SSN only}} Capturing only the dynamic user behavior via sequential information in lists does not give the optimal performance. This shows that users have a general preference towards items, which SSN does not consider.
        
        \subsubsection*{\textbf{MGNN and SSN (UHGNN removed)}} We model the collaborative signals in multi-hop homogeneous (same entity) neighbors from the available interactions through MGNN, along with the sequential information provided by SSN. The additional information leads to a considerable increase in the performance across all the datasets, particularly in Goodreads. This further shows the efficacy of GNNs for recommendations.
        
        \subsubsection*{\textbf{MGNN, UHGNN and SSN (HyperTeNet)}} As can be seen from \autoref{fig:ablation_study}, when all the three sub-models are present, we consistently get the best performance, even beating the current state-of-the-art models in the process. Due to UHGNN, the embeddings also capture the \textit{ternary relationship} among the heterogeneous entities (users, items, and lists), demonstrating the importance of capturing it, which went overlooked in the previous works.
    
    \vspace{0.3cm}
    \noindent
    \textbf{Effect of using MGNN embeddings.} \autoref{fig:modeling_choice} demonstrates the effectiveness of using MGNN embeddings ($p'_u, q'_i, t'_l$) as inputs to UHGNN and SSN, instead of using the plain embeddings ($p_u, q_i, t_l$) obtained directly from the embedding lookups. The better metric scores associated with MGNN embeddings can be attributed to the higher-order graph structure being appreciated with them, as against using the plain embeddings.
    
    \subsection{RQ3: Analyzing Hyperparameters}
    \begin{figure}[htp]
        \centering
        \begin{subfigure}{0.48\columnwidth} 
            \includegraphics[height=100pt, width=\columnwidth]{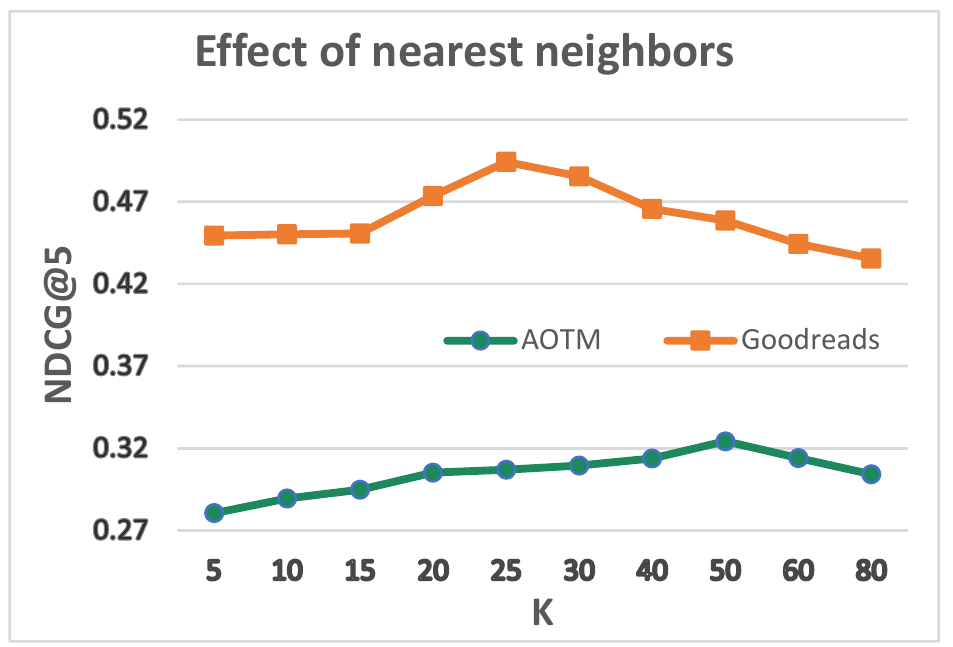}
            \caption{Number of nearest neighbors} 
            \label{fig:hyperparameter_study_a}
        \end{subfigure} 
        \hfill
        \begin{subfigure}{0.45\columnwidth} 
            \includegraphics[height=100pt, width=\columnwidth]{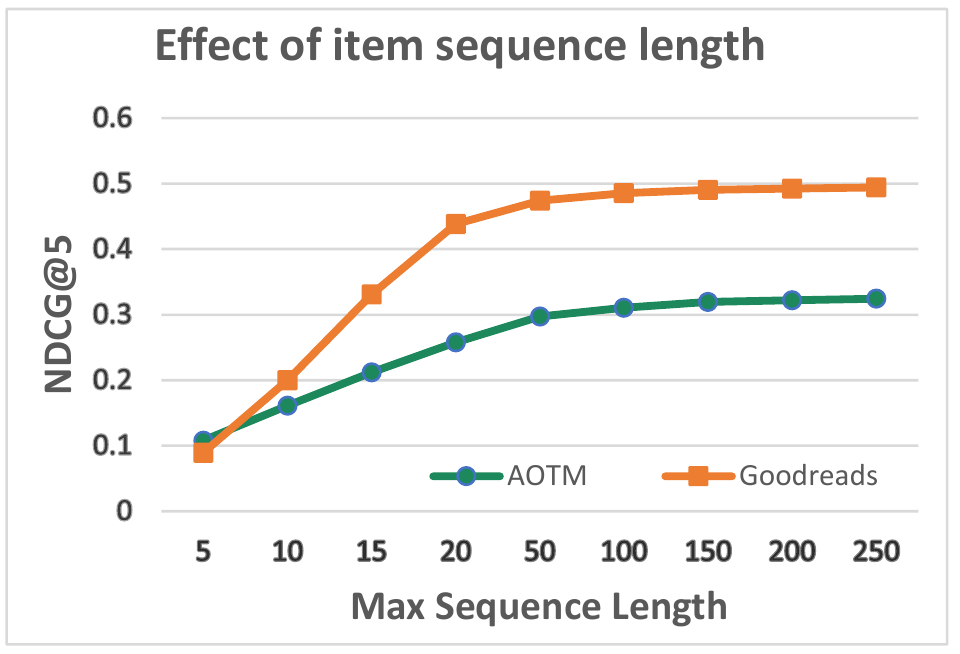}\hfill
            \caption{Maximum number of items} 
            \label{fig:hyperparameter_study_b}
        \end{subfigure} 
        \hfill
        \caption{Impact of hyperparameters on NDCG@5 on AOTM and Goodreads datasets.}
        \label{fig:hyperparameter_study}
    \end{figure}
    \autoref{fig:hyperparameter_study_a} demonstrates the effect of the number of neighbors, $K$, in the K-NN graph. In all four datasets, we observe a mountain-like structure. For Goodreads, the peak occurs at $K = 25$, while for the rest, the peak occurs at $K = 50$. \autoref{fig:hyperparameter_study_b} shows the impact of fixing a maximum number of items in a sequence. The observations suggest that the larger the sequence length, the better is the performance of the model. The scores, however, saturate for larger values ($> 200$). Conversely, by fixing shorter lengths of lists, the performance degrades quite significantly. This observation aligns with the observations of CAR \cite{he2020consistency} that users have a general preference that is always maintained, determining the items that are added next to the list. This further suggests that HyperTeNet is indeed able to capture this general user preference for personalized list continuation.

\section{Conclusion}
In this work, we proposed HyperTeNet -- a self-attention-hypergraph and Transformer-based architecture for the personalized list continuation task. HyperTeNet understands the multi-hop relationships among the entities of the same type and learns ternary relationships among users, items, and lists by leveraging graph convolutions on a 3-uniform hypergraph, a crucial idea behind the success of the current approach. Further, the use of Transformers helps the proposed model understand the dynamic preferences of users, thereby helping it to identify the next item(s) to be included in the list(s). Our experiments on four real-world datasets demonstrate that HyperTeNet outperforms the other state-of-the-art models. At the same time, the ablation studies on different constituent networks show the efficacy of each network.

The proposed model is generic as it is easy to extend it to capture $k$-ary relationships among $k$ entity types, coming from different modalities, by constructing a $k$-uniform hypergraph. This makes some constituents of HyperTeNet beneficial for tasks other than recommendations too. In the future, we plan to extend these ideas to entities coming from different modalities such as images, audio, and videos for recommendations.

% We use a training procedure that alternates between minimizing loss corresponding to graph and hypergraph neural networks and transformer-based sequence networks to learn personalized playlist continuation. 
% \section*{Acknowledgment}
\bibliographystyle{./IEEEtran}
\bibliography{./tenet}
\end{document}